\documentclass[lettersize,journal]{IEEEtran}
\usepackage{amsmath,amsfonts}
\usepackage{algorithmic}
\usepackage{algorithm}
\usepackage{array}
\usepackage[caption=false,font=normalsize,labelfont=sf,textfont=sf]{subfig}
\usepackage{textcomp}
\usepackage{stfloats}
\usepackage{tabularx}
\usepackage{tabulary}
\usepackage{booktabs} 
\usepackage{url}
\usepackage{verbatim}
\usepackage{graphicx}
\usepackage{cite}
\usepackage{longtable}
\usepackage{comment}
\usepackage{supertabular}
\usepackage{multirow}
\usepackage{hyperref}
\usepackage[normalem]{ulem}
\useunder{\uline}{\ul}{}
\usepackage[table,xcdraw]{xcolor}

\newcommand{\attribute}{\textit}

\begin{document}

\title{Dynamic loss balancing and
sequential enhancement for road-safety assessment\\
and traffic scene classification}

\author{Marin Kačan, Marko Ševrović, Siniša Šegvić
\thanks{This work has been co-funded by the Connecting Europe Facility of the European Union (project SLAIN: Saving Lives Assessing and Improving TEN-T Road Network Safety), Croatian Science Foundation (grant IP-2020-02-5851 ADEPT), the European Regional Development Fund (project KK.01.1.1.01.0009 DATACROSS), and NVIDIA Academic Hardware Grant Program.}
\thanks{M. Kačan and S. Šegvić are with University of Zagreb Faculty of Electrical Engineering and Computing (e-mail: {marin.kacan, sinisa.segvic}@fer.hr).}
\thanks{M. Ševrović is with University of Zagreb Faculty of Transport and Traffic Sciences (email: marko.sevrovic@fpz.unizg.hr)}
}

\markboth{Journal of \LaTeX\ Class Files,~Vol.~14, No.~8, August~2021}%
{Shell \MakeLowercase{\textit{et al.}}: A Sample Article Using IEEEtran.cls for IEEE Journals}

\IEEEpubid{0000--0000/00\$00.00~\copyright~2021 IEEE}

\maketitle

\begin{abstract}

Road-safety inspection is
an indispensable instrument
for reducing road-accident fatalities 
related to road infrastructure.
Recent work formalizes
the assessment procedure in terms of
carefully selected risk factors that are also known as road-safety attributes.
In current practice, these attributes
are manually annotated
in geo-referenced monocular video
for each road segment.
We propose to reduce 
dependency on tedious human labor
by automating attribute collection 
through a two-stage deep learning approach.
The first stage recognizes more than forty road-safety attributes
by observing a local spatio-temporal context.
Our design leverages an efficient convolutional pipeline,
which benefits from pre-training
on semantic segmentation of street scenes.
The second stage enhances predictions
through sequential integration
across a larger temporal window.
Our design leverages per-attribute instances of a lightweight recurrent architecture.
Both stages alleviate extreme class imbalance
by incorporating a multi-task variant
of recall-based dynamic loss weighting.
We perform experiments on the novel iRAP-BH dataset,
which involves fully labeled geo-referenced video
along 2,300 km of public roads in Bosnia and Herzegovina.
Moreover, we evaluate our approach against the related work
on three road-scene classification datasets from the literature: Honda Scenes, FM3m, and BDD100k.
Experimental evaluation confirms the value
of our contributions on all three datasets.

\end{abstract}



\section{Introduction}
\label{sec:introduction}

\IEEEPARstart{R}{oad} accidents are a significant public health problem that causes more than 1.3 million deaths every year \cite{5}.
With its Global Plan for the Decade of Action for Road Safety \cite{un_global_plan}, the UN has committed to halving the number of road traffic deaths and injuries by the end of this decade.
The holistic Safe System Approach \cite{safe_system_01,safe_system_02} identifies road infrastructure as one of the five pillars of traffic safety \cite{safe_system_03,li24pone}.

Many earlier approaches to road-safety assessment estimate high-risk road sections (black spots) from historical data \cite{blackspot_meta}.
These approaches are reactive in nature since they require accidents to occur for a section to be deemed dangerous \cite{blackspots_02,blackspots_03}.
In contrast, a proactive approach to road safety formalizes the safety in terms of road-infrastructure attributes.
Besides being able to operate when historical data are absent, this approach also provides an insight into specific requirements  to improve the safety rating.
While road-safety attributes can be assessed via on-site surveys \cite{6}, off-line assessments are usually more practical and cost-effective \cite{7,8}.
Currently, these assessments are performed manually by trained operators \cite{3}.
Automating this time-consuming process is a step toward reduced costs and increased scalability.

\IEEEpubidadjcol

This work presents a two-stage visual recognition approach for assessing road-safety attributes in monocular video.
Our models 
operate on 
street-level imagery by complementing 
local recognition 
and sequential enhancement with dynamic loss weighting.
We implement 
local recognition 
as a multi-task \cite{bengio13pami} model with shared latent representation
and per-attribute classification heads.
We facilitate high-quality latent representation 
by pre-training on a large street-level semantic segmentation dataset \cite{vistas}.
Our sequential model enhances 
local predictions
by providing access  
to a larger temporal context.
It consists of 
per-attribute bidirectional LSTM models 
in order to account for 
attribute-specific temporal behavior.
We address the extreme imbalance of training data by proposing 
a multi-task variant of dynamic loss weighting 
with respect to per-class recall \cite{recall_loss_01}.
Such practice benefits both 
local recognition 
and sequential enhancement.

We evaluate our approach on a novel dataset for road-safety assessment, which we denote as iRAP-BH.
We have acquired the dataset along 2,300km of public roads in Bosnia and Herzegovina.
It covers 230,000 10-meter road segments annotated with all 52 iRAP attributes \cite{3}.

The proposed approach has no handcrafted attribute-specific rules.
Hence, it could contribute to different road-safety programmes and visual event recognition problems.
We demonstrate this through performance evaluation on three public road-driving classification datasets: Honda Scenes \cite{narayanan19icra}, FM3m \cite{sikiric20tits}, and BDD100k \cite{yu20cvpr}. 
Experiments reveal that our approach outperforms the state of the art on all tasks of Honda Scenes.
The greatest improvements occur on the Road place task which 
contains temporal annotations and highly imbalanced classes. 
Our models also perform very well on FM3m and BDD100k,
even though the former is fairly well balanced and both are unsuitable for multi-frame recognition.

The contributions of this paper are as follows.
Firstly, building on our preliminary research \cite{kacan20itsc}, we enhance attribute classification performance through pre-training on the task of semantic segmentation. This 
favours extraction of localized semantic features 
from street scenes.
Empirical evidence shows a clear advantage
with respect to other forms of pre-training, 
and highlights the importance of 
nuanced visual representations 
for road-attribute recognition.
Secondly, we advance beyond 
our preliminary work \cite{kacan20itsc} 
by addressing all road-safety attributes 
that can be visually assessed, 
as well as all of their fine-grained classes.
Thirdly, we mitigate class imbalance through multi-task dynamic loss weighting. 
Our policy dynamically adjusts class weights with respect to the corresponding recall scores. 
Furthermore, it normalizes per-attribute loss components 
in order to prevent undesired cross-attribute interference.
This enables effective joint learning across many attributes, even in the presence of extreme class imbalances.
Fourthly, we enhance our 
local predictions with 
per-attribute sequential models. 
These models can learn 
improved attribute-specific temporal patterns
by observing a wider temporal context.
We provide thorough ablation studies which demonstrate the impact of our contributions.

The remainder of this paper is structured as follows. Section \ref{sec:related_work} surveys the related work in the areas of road-safety assessment and road-driving image classification. Section \ref{sec:road-safety_attributes} introduces and analyzes a widely used collection of road-safety attributes. We describe our proposed approach and its technical contributions in Section \ref{sec:model}. Section \ref{sec:datasets} outlines the datasets utilized in our experiments. 
Section \ref{sec:experiments} describes the conducted experiments and analyzes the results. Finally, Section \ref{sec:conclusion} draws conclusions from our work, highlights the significance of our contributions and considers their implications for future research.

\section{Related work}
\label{sec:related_work}

\subsection{Road infrastructure safety}

Traditional approaches to road-infrastructure safety assessment 
are based on historical accident statistics.
Time, location, severity, type, and other data \cite{chatterjee19sage} about previous crashes can be used to identify black spots \cite{cui22jtte},  create crash-risk maps \cite{dimitrijevic22as}, and design crash-prediction models \cite{gaweesh19aap}.
These approaches can capture complex and non-trivial risk factors that history-agnostic approaches might overlook \cite{blackspots_03}.
However, these approaches are reactive since they require accidents to occur for road sections to be assessed as unsafe \cite{itf22oecd}. 
Moreover, historical estimation tends to produce high-variance predictions due to the sparsity of accidents \cite{he21iccv}.

Proactive approaches to safe road infrastructure 
rely on periodic inspections of static features.
The International Road Assessment Programme (iRAP) Star Rating \cite{7} 
is an internationally accepted standard for assessing and improving the safety of road infrastructure.
Empirical evidence shows that it can significantly reduce the number of traffic fatalities and serious injuries by identifying high-risk roads across many countries \cite{li24pone,8}.
The iRAP Rating assesses the in-built safety of road segments according to 52 attributes \cite{3} 
that we review in Section \ref{sec:road-safety_attributes}.

\subsection{Computer vision for road-infrastructure assessment}

Computer vision has been applied to detect and recognize various elements of road infrastructure 
such as traffic signs \cite{tabernik20tits,segvic14mva}, road surface markings \cite{zou20tvt,zadrija18,mccall06tits}, 
fleet management attributes \cite{sikiric20tits}, 
etc.
Outputs from 
detection and segmentation models \cite{bulo18cvpr,cheng21nips,kreso20tits,orsic} have been 
utilized as intermediate results to recognize certain road-safety attributes 
with rule-based systems \cite{9,zohaib18ivcnz,sanjeewani20ivcnz}.
For instance, Yi et al.~\cite{yi21icbd} leverage active learning to detect road-safety elements such as guardrails and utility poles.
Ni et al. \cite{ni22tim} use an object detection network to extract local features to augment a global feature-extracting convolutional network for road-place classification.
These approaches can benefit from more advanced data augmentation procedures such as CutMix \cite{yun19iccv} or Mosaic \cite{bochkovskiy20arxiv}, which have been beneficial for detection performance in road and railway environments \cite{mauri22jrtip}.

In contrast, 
our 
local recognition pipeline 
aims to recognize road-safety attributes
directly from input data, in an end-to-end image-wide manner \cite{kacan20itsc}.
This avoids error accumulation
and produces better latent representations. 
In this vein, the authors of the Honda Scenes dataset \cite{narayanan19icra} 
present a baseline approach for infrastructure-related 
event detection in street video.
They pre-train 
the ResNet-50 backbone on Places365
and leverage a frozen semantic segmentation model
to ignore traffic participants.
The pipeline concludes by 
recurrent processing
of frozen convolutional features
and standard softmax classification. 
Unlike their recurrent task-agnostic temporal region proposals 
and subsequent classification into events, 
our method enhances existing convolutional predictions 
for specific tasks using recurrent processing.
Context MTL \cite{lee20iv}
addresses recognition on Honda Scenes
with a multi-task architecture
that is related to our work \cite{kacan20itsc}.
They regularize the loss with a lower bound
of mutual information between the input and the latent-space features 
according to the Jensen-Shannon divergence \cite{hjelm19iclr}. 
Multi-Task Attention Network (MTAN) \cite{liu19cvpr} uses a shared WideResNet backbone to generate task-specific attention masks for dense prediction. It dynamically selects relevant features from the shared global feature map for each task. The network is trained using a combined loss function where task-specific losses are weighted according to their current rate of change. 
None of the described approaches address class imbalance with 
multi-task dynamic loss weighting, nor use 
semantic-segmentation pre-training.


\subsection{Learning on imbalanced data}

Class imbalance refers to non-uniform class proportions within training data \cite{herrera18springer}. 
Single-task setups 
often mitigate imbalance with 
data-level strategies such as oversampling rare classes and undersampling frequent ones \cite{bevandic24ijcv,bulo18cvpr}. 
However, these methods are less feasible in multi-task scenarios \cite{lee20iv,liu19cvpr} where tasks are uncorrelated and 
non-uniformly imbalanced. 
In such cases, an image marked as rare in one task might simultaneously be frequent in another, 
which complicates the direct application of oversampling. 
A possible alternative 
would be to construct dedicated 
oversampled datasets for each task 
and then train the multi-task model 
by cyclically optimizing tasks 
in a round-robin fashion. 
However, our preliminary experiments 
indicate that round-robin training 
underperforms with respect to 
standard training 
by a wide margin. 
Consequently, we choose to address 
multi-task class imbalance 
by a custom loss weighting approach. 

Several algorithmic-level and cost-based approaches try to adapt learning algorithms by assigning larger loss weights to misclassified examples of underrepresented classes \cite{kreso16gcpr,inv_ce_loss_01,inv_ce_loss_02,balanced}.
However, simple implementations of inverse-frequency loss-weighting improve recall at the expense of precision 
\cite{recall_loss_01}.
Tian et al.\cite{recall_loss_01} address this issue by introducing dynamic assignment of class weights.
During training, the class weights are dynamically set according to the current false negative rate of the corresponding class. 
This prevents the classes that achieve high recall from suffering excessive false positives. 
Unfortunately, this approach 
is not directly applicable 
in multi-task setups 
as we explain in Section \ref{subsec:loss}.

\subsection{Domain shift}

Robustness to domain shifts with regards to environment or driving conditions 
is important for real-world visual recognition systems \cite{sakaridis21iccv,bevandic22ivc,zendel18eccv}.
Domain adaptation approaches 
have successfully been employed for cross-domain traffic scene recognition.
Saffari et al.\cite{saffari23tetci} 
perform traffic scene classification under varying weather conditions 
by combining 
a generative domain-invariant feature extractor, with nonlinear task-relevant dictionary learning. 
Di et al.\cite{di21tits} tackle semantic segmentation of rainy night-time scenes by introducing a near-scene semantic adaptation approach that leverages the 
corresponding daytime images. 
They minimize domain shift on the feature representation level and subsequently align the segmentation output space of the pre-trained daytime model with the rainy night-time domain.
They evaluate the approach on a dataset with paired daytime and rainy night-time images. 

Our road-safety attribute dataset iRAP-BH is not applicable for domain adaptation experiments, since it does not contain driving conditions annotations nor images from different countries. 
We aim to improve within-domain 
robustness through 
semantic segmentation pre-training, 
loss balancing and 
sequential enhancement.

\subsection{Recurrent models for video recognition}

Long Short-Term Memory (LSTM) networks \cite{hochreiter97neco} have been used for video classification and action recognition \cite{ng15cvpr}.
They have also been utilized to enhance existing sequential predictions in speech recognition \cite{tu17apsipa} and rainfall regression \cite{kratzert18hess}. 

Inspired by temporal region proposal approaches \cite{xu17iccv,chao18cvpr}, Narayanan et al. \cite{narayanan19icra} use an LSTM-based 
architecture to perform decoupled event proposal and traffic scene classification. 
Their two-stage approach generates task-agnostic event proposals as video intervals and subsequently 
classifies those into events from spatio-temporally 
pooled features.
Since these two stages are decoupled, the second stage is treated as a single-task multi-class learning problem.
Trabelsi et al. \cite{trabelsi22ieeeaccess} extend the LSTM network with multi-head attention and combine it with a convolutional neural network to effectively capture and interpret complex dynamics of driver behaviour from traffic scenes.
In contrast, we use recurrent processing to enhance 
existing convolutional predictions for a specific task, rather than for task-agnostic feature aggregation.

\section{Road-safety attributes}
\label{sec:road-safety_attributes}

The 
iRAP Star Rating quantifies the overall protection that road infrastructure provides to the four most common road user types \cite{3}.
The assessment procedure targets 
categorical values of 
a carefully selected set of 52 attributes
related to road-infrastructure elements and roadside objects 
within the corresponding 100-meter or 10-meter road segment.
Each 
attribute 
assumes a class from the attribute-specific taxonomy.
The number of classes varies across attributes.
Attributes with the most classes are those that encode the speed limit (21 classes), roadside severity (17 classes), and intersection type (16 classes).
On the other hand, there are 11 binary attributes.
Hence, we formulate attribute recognition as separate multi-class classification problems.
We had to discard four attributes that assume only one class throughout our whole dataset: \attribute{Shoulder rumble strips}, \attribute{Centre line rumble strips}, \attribute{Motorcycle facility}, and \attribute{Pedestrian fencing}. 
We also had to discard five attributes that are not suitable for a visual recognition setup. 
These include the four speed limit attributes and \attribute{Intersecting road volume}. 
The latter captures the average daily number of vehicles that pass through a road segment from an intersecting road.
This information can be estimated from 
road counters 
or aerial imagery. 
Consequently, our experiments address 43 iRAP attributes.

\subsection{iRAP Attribute Groups}
We provide a short overview of the seven attribute groups defined by the iRAP standard \cite{3}.
More details can be found in our preliminary work \cite{kacan20itsc}.

\attribute{Road and context attributes} (1 attribute) contain the attribute Carriageway label along with twelve attributes related to data acquisition and annotation metadata.

\attribute{Observed flow attributes} (5 attributes) record the flow of motorcycles, bicycles, and pedestrians 
through the segment. 

\attribute{Speed limit attributes} (5 attributes) record  
the 
speed limits (4 attributes) and 
speed-reducing infrastructure such as speed bumps.

\attribute{Mid-block attributes} (16 attributes) focus on the road's intrinsic features rather than its surroundings. 
The attribute Median type stands out as particularly challenging, as it requires distinguishing among 15 kinds of physical separators or median markings.

\attribute{Roadside attributes} (7 attributes) involve dual-side (passenger and driver) attributes 
that asses the risk of roadside features. 
The attribute Roadside Severity captures the most hazardous roadside object based on its type and proximity to the road.
The ground truths for this attribute are assigned according to the priority table \cite{3} 
that ranks object and distance combinations according to the risk level.

\attribute{Intersection attributes} (5 attributes) capture various intersection characteristics. 
Most notably, the attribute Intersection type has 16 classes that cover different combinations of intersecting roads, signalization, and special features like roundabouts and railway crossings.

\attribute{Vulnerable road-user facilities and land use attributes} (13 attributes) detail the presence of pedestrian, cyclist, and motorcyclist amenities, as well as capturing the characteristics of the surrounding area (e.g.\ Area type, Land use, School zone).

\subsection{Attribute analysis}
\label{subsec:imbalance}
We provide a conceptual and empirical analysis of various aspects of iRAP attributes in our dataset.

\subsubsection{Class imbalance}
Many attributes in our dataset suffer from class imbalance, 
which occurs when there is a significant disproportion among the number of examples belonging to 
different classes. 
Class imbalance can hamper the performance of accuracy-oriented classifiers, resulting in the minority classes being ignored \cite{herrera18springer}. 
We thoroughly study and address this issue through improved loss functions, training procedures and macro-F1 evaluation, as described in \ref{subsec:loss} and \ref{subsec:metrics}.

\subsubsection{Non-orthogonal design}
Some iRAP attributes capture multiple features that seem orthogonal to each other, resulting in classes that are (nearly) cartesian products of different values of those features.
For example, the attribute \attribute{Skid resistance} is supposed to capture the skidding resistance and the texture depth of the road surface.
It covers two dimensions: whether the surface grip is low (“poor”), medium, or adequate; and whether the road is sealed or unsealed.
This results in the attribute having the following 5 classes: \attribute{Unsealed - poor}, \attribute{Unsealed - adequate}, \attribute{Sealed - poor}, \attribute{Sealed - medium}, \attribute{Sealed - adequate}.
The attribute might have been divided into two attributes: \attribute{Sealed road} (true/false) and \attribute{Surface grip} (poor/medium/adequate).
An orthogonal formulation of these concepts would mitigate the issue of class imbalance since there would be fewer classes with more examples.

\subsubsection{Fine-grained and visually similar classes}
Some attributes have very fine-grained classes that can be visually very similar.
For instance, the attribute \attribute{Roadside severity} contains classes that cover different types of safety barriers (metal, concrete, wire, motorcycle friendly) 
as well as a separate class for semi-rigid structures, such as various fences.
These classes are well defined, but the many options make it easy to miss the correct answer and also exacerbate class imbalance by spreading the already infrequent examples over many classes. 

In general, it might be advantageous to design attribute sets with fewer and more general classes. 
Such a decision would trade off some precision and specificity for improved recognition quality.

\subsection{Temporal behaviour}
\label{subsec:analysis_sequential}
The videos in our dataset cover long road sections.
A road section is composed of a sequence of successive 10-meter segments.
We identify several distinct patterns of temporal 
attribute behavior along sequences.
According to the iRAP standard, some attributes have a default "negative" class \cite{3}.
These attributes are usually concerned with capturing countable occurrences of various infrastructure elements, such as intersections or pedestrian crossings.
The default class in these attributes is \attribute{None}, 
while the "positive" classes 
correspond to concrete realizations of that attribute 
(e.g. \attribute{3-leg intersection}). 
The iRAP standard mandates that any occurrence of a positive class should only be annotated in the segment closest to its occurrence.
All other neighboring segments are annotated as the negative class.
We call such attributes "single-peak" attributes.

Let us consider a segment that contains an occurrence of a positive class (a peak) and a few neighboring segments that appear immediately before it.
The visual features that a model might use to recognize such an attribute 
will typically be present in the neighboring segments as well. 
For example, an intersection gradually becomes more and more visible in the segments leading up to the peak segment.
A model that predicts an intersection in a neighboring segment is not entirely wrong, and it might be hard for it to discern which exact segment is the peak one.

The attribute \attribute{Street lighting} has an 
especially peculiar temporal behaviour. 
It is treated as a single-peak attribute when a single light post appears in isolation.
On the other hand, for a sequence of light posts, street lighting should be recognized 
as present 
in all segments from the first to the last light post.
Since the light posts in such sequences can be up to 100 meters apart, it may be difficult for a vision-based model to differentiate between single occurrences and sequences of street lights.

There is another subset of attributes that is opposite in nature to single-peak attributes.
We call these attributes "smooth" since their classes rarely change and generally do not oscillate.
They describe larger areas, environments, zones, or infrastructure features that are 
likely to remain unchanged in consecutive segments. 
Examples of such attributes include \attribute{Area type}, \attribute{Road delineation}, \attribute{Carriageway label}, etc.

Figure \ref{fig:LSTMExamples} shows examples of four iRAP attributes.
It presents sequences of five frames from consecutive 10-meter segments, along with the ground truth labels and model predictions for the corresponding attribute.
Only the third segment of row 1 is annotated with the positive class 
of the single-peak attribute \attribute{Intersection type}. 
Conversely, the two smooth attributes maintain constant ground truth label 
throughout rows 3 and 4. 
The ground truth label of \attribute{Street lighting} also does 
not change even though the discriminative visual features are not visible in certain segments.

\subsection{Motivation for sequential enhancement}
This subsection analyzes class co-occurrence along consecutive segments of iRAP-BH. 
For a given attribute \textit{A} and a given pair of segments \textit{t} and \textit{(t+1)}, the pair of corresponding classes $(\mathrm{c_{A,t}},\mathrm{c_{A,t+1}})$ constitutes a single co-occurrence.
For an attribute with \textit{n} classes, we can construct an $n \times n$ co-occurrence matrix where the element $(i,j)$ corresponds to the number of occurrences 
where $\mathrm{c_{A,t}} = i$ and $\mathrm{c_{A,t+1}} = j$.
We build two such matrices for each attribute: one using the ground truth labels and the other using predictions produced by our 
local recognition pipeline 
from Figure \ref{fig:cnn_single_frame}.
For single-peak attributes, the only possible ground truth transitions go from the default class to a positive class and back.
Thus, their ground truth matrices will have non-zero elements only in the row and the column of the default class.
For smooth attributes, most consecutive segments belong to the same class.
Thus, 
the diagonal elements of their ground truth matrices will be significantly larger than off-diagonal elements.

Our analysis reveals consistent discrepancies between ground truth and 
local 
prediction matrices for these two groups of attributes.
In the 
local prediction matrices, single-peak attributes have 
many non-zero diagonal values.
These 
discrepancies occur when 
the model assigns the same positive class to two consecutive segments that are visually very similar (e.g. two segments inside an intersection).
This is a reasonable error, but it shows that the convolutional recognition model fails to learn the single-peak annotation convention.
For smooth attributes, significantly larger values of off-diagonal elements were observed in prediction matrices compared to ground truth matrices.
This reveals the presence of spurious class transitions in 
local 
predictions for consecutive segments, which implies that our local model fails to learn the inertia of smooth attributes.

This result motivates us 
to extend the 
local recognition pipeline 
with per-attribute sequential enhancement models that learn temporal behavior patterns 
without requiring costly backpropagation through hundreds of video frames.
The technical details of this component 
are described in \ref{subsec:sequential}.

\section{Multi-task recognition of road-safety attributes in video}
\label{sec:model}

Our approach performs multi-task 
recognition of road-safety attributes through 
two stages: 
local recognition (section \ref{subsec:conv}) 
and sequential enhancement (section \ref{subsec:sequential}). 
Both stages alleviate class imbalance 
by dynamic loss weighting according to our 
multi-task recall analysis (section \ref{subsec:loss}).

\subsection{Recognition in the local spatio-temporal context}
\label{subsec:conv}

\begin{figure}[ht]
    \centering
    \includegraphics[width=\linewidth]
      {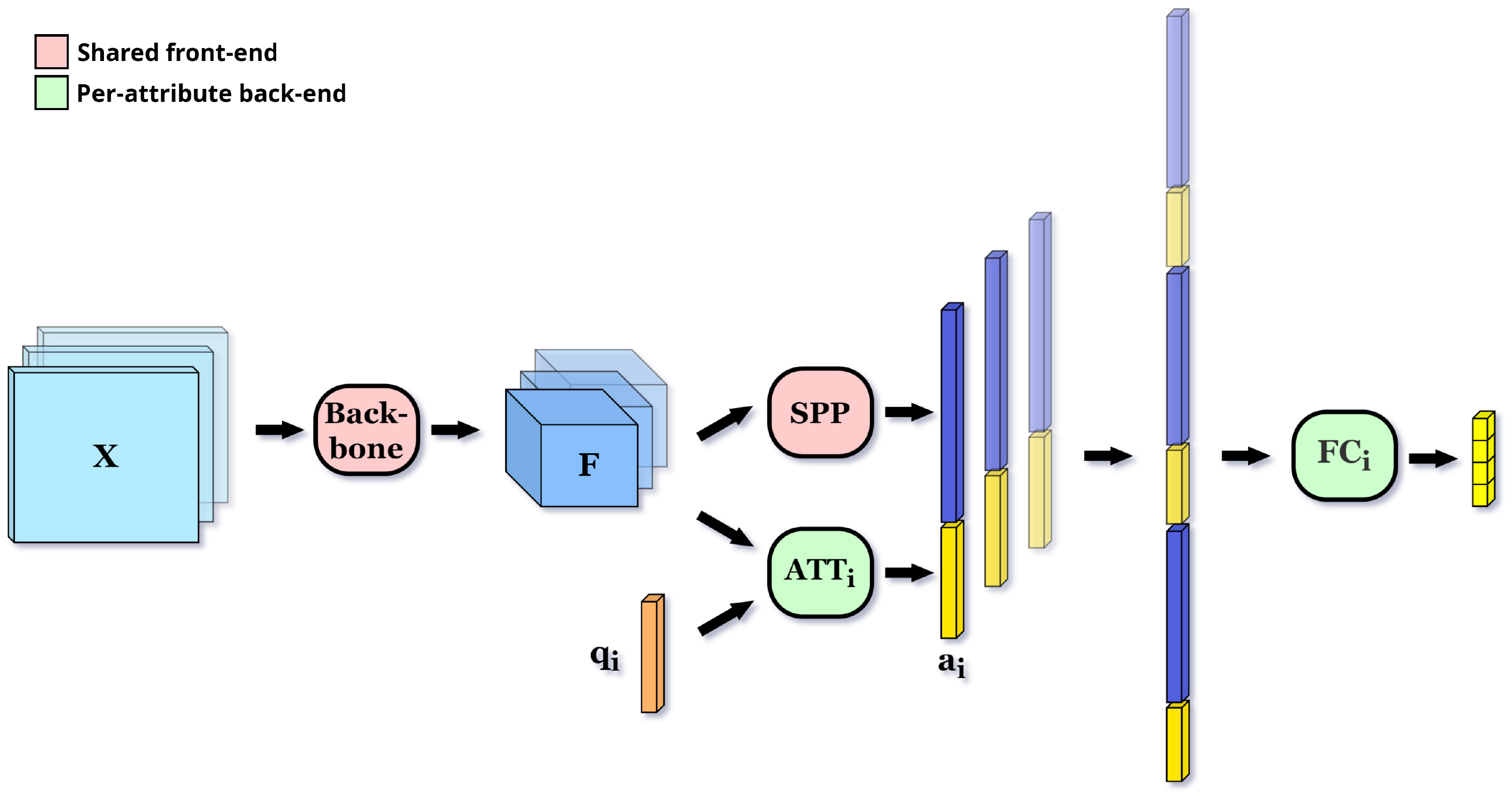}
    \caption{
Our multi-frame 
local recognition pipeline 
recognizes road-safety attributes 
in multi-frame input $\mathbf{X}$.
Data tensors are represented as cuboids, while processing modules are shown as rounded rectangles. 
The shared front-end (red) 
maps each input frame into 
convolutional features ($\mathbf{F}$) 
that are subsequently 
pooled by the $\mathbf{SPP}$ module.
Attribute-specific back-ends (green) 
produce attention pools (yellow) 
and concatenate them 
with the shared spatial pools (blue).
Fully-connected layers ($\textrm{FC}_i$) map the 
concatenated descriptors into 
attribute-specific logits.
     }
    \label{fig:cnn_single_frame}
\end{figure}

Figure \ref{fig:cnn_single_frame} presents our convolutional architecture  
for multi-task visual recognition of road-safety attributes in street-level imagery.
The architecture consists of a shared front-end
and attribute-specific back-ends.
The front-end starts with the ResNet-18 backbone
which we pre-train for semantic segmentation
on the Vistas dataset \cite{vistas,orsic21pr}. 
The resulting features are subjected
to Spatial Pyramid Pooling (SPP)
with grid dimensions 6, 3, 2 and 1 \cite{spp}.
The SPP module captures information 
at different scales and produces 
a shared fixed-size image-wide representation.

Each of the 43 attribute-specific back-ends
starts with attention pooling $ATT_i$ \cite{kacan20itsc}
with respect to the learned attribute-specific query $q_i$.
The resulting representation is concatenated
with the shared SPP features 
into the single-frame descriptor 
that is 
fed to the corresponding prediction head $P(A_i|\mathbf{x})$.

This architecture is easily extended  
for multi-frame input.
In that case, 
several single-frame attribute descriptors 
are concatenated into 
the multi-frame attribute descriptor.
Per-attribute back-ends remain the same
as in the single-frame case. 
Note that the number of input frames 
can not be arbitrarily large in order to avoid 
memory exhaustion during training.
All our multi-frame models
produce predictions for segment T 
by observing the middle frames 
of segments T, T-1, and T-4.
Each per-attribute prediction head is subject 
to the corresponding cross-entropy loss. 
Following the multi-task learning paradigm \cite{bengio13pami}, 
the total loss is the mean 
of all per-attribute losses.

\subsection{Dynamic loss weighting for multi-task learning}
\label{subsec:loss}

We wish to alleviate multi-task class imbalance (cf. \ref{subsec:imbalance}) by increasing the influence of rare classes on the training objective.
We denote our training set with $\left\{x_{n}, y_{n}\right\}$, where $x_{n} \in R^{d}$, $y_{n} \in\{1, \ldots, C\}$, and $n \in\{1, \ldots, N\}$.
Let $P_{n}^{c} = P(Y=c|x_n)$ denote the predicted posterior of class $c$ for input $x_{n}$. The standard cross-entropy can be interpreted as negative logarithm of the geometric mean of the correct class posterior 
$\overline{P}=\left(\prod_{n=1}^{N} P_{n}^{y_{n}}\right)^{1 / N}$ 
over all examples \cite{recall_loss_01}:

\begin{align}
\mathrm{CE} 
 &=-\frac{1}{N} \sum_{n=1}^{N} 
     \ln P_{n}^{y_{n}}
   =-\frac{1}{N} \ln \left(\prod_{n=1}^{N} P_{n}^{y_{n}}\right) 
 \nonumber \\
 &=-\ln \left(\prod_{n=1}^{N} P_{n}^{y_{n}}\right)^{\frac{1}{N}}
  =-\ln \overline{P}
\end{align}
This equation can also be expressed as a weighted sum of per-class geometric means $\overline{P^c}$:
\begin{align}
 \mathrm{CE} 
 &=-\frac{1}{N} \sum_{c=1}^{C} \sum_{n: y_{n}=c} 
   \ln P_{n}^{c}
  =-\sum_{c=1}^{C} \frac{1}{N} 
   \ln \left(\prod_{n: y_{n}=c} P_{n}^{c}\right) 
 \nonumber \\
 &=-\sum_{c=1}^{C} \frac{N_{c}}{N} 
   \ln \left(\prod_{n: y_{n}=c}
   \!
   P_{n}^{c}\right)^{\frac{1}{N_{c}}}
   \!\!\!
  =-\sum_{c=1}^{C} \frac{N_{c}}{N} 
   \ln \overline{P^{c}}
\label{eq_02}
\end{align}
The range $\left\{n: y_{n}=c\right\}$ denotes examples of class $c$. 
The symbol ${\overline{P^{c}}=\left(\prod_{n: y_{n}=c} P_{n}^{y_{n}}\right)^{1 / N_{c}}}$ denotes the geometric mean posterior of the correct class in samples that belong to class $c$. 
The equation shows that standard cross-entropy maximizes the weighted arithmetic mean of per-class geometric mean posteriors with the weights being the relative class frequencies $N_{c}/N$.

If we want each class to have the same contribution to the loss, we can assign a weight to each class that is the inverse of its relative frequency: $w_{c} = N/N_{c}$. This yields the inverse-frequency-weighted cross-entropy loss \cite{inv_ce_loss_01,inv_ce_loss_02}:
\begin{align}
\mathrm{CE^{IFW}} 
 &=-\frac{1}{N} \sum_{n=1}^{N} w_{y_{n}} 
   \ln P_{n}^{y_{n}}
  =-\frac{1}{N} \sum_{c=1}^{C} w_{c}\sum_{n: y_{n}=c} 
   \ln P_{n}^{y_{n}} \nonumber \\
 &=-\sum_{c=1}^{C} \frac{1}{N} w_{c} \frac{N_{c}}{N} 
   \ln \overline{P^{c}}
  =-\sum_{c=1}^{C} \frac{1}{N} \ln \overline{P^{c}}
\end{align}

The standard cross-entropy does not take into account the distribution of the posterior over incorrect classes.
In that sense, cross-entropy can be viewed as measuring 
the extent to which a particular prediction is a false negative 
while ignoring the false positives. 
Thus, assigning a large weight to a particu lar class 
might also increase the incidence of false positives for that class.
This analysis has been confirmed empirically \cite{recall_loss_01}: 
increasing the class weight indeed decreases the precision of predictions for that class.

These observations suggest that 
placing a large weight on a rare class
that already achieves high recall 
is likely to decrease precision
with little gain in recall.
This can be prevented
by adapting the loss weights 
with respect to the model performance 
in terms of per-class recall \cite{recall_loss_01}.
Let $R_{c,t}$ denote the validation recall of class $c$ after epoch $t-1$.
Then, recall-balanced class weights $w^R_{c,t}$ 
can be expressed as \cite{recall_loss_01}: 

\begin{equation}
w_{c,t}^{R} 
=\frac{N}{N_{c}}\left(1-R_{c,t}\right) + \epsilon
\label{eq:recall_loss}
\end{equation}
When recall is close to zero, the weight (\ref{eq:recall_loss}) approaches the inverse relative frequency.
As the recall of a class increases, its weight diminishes. 
We add $\epsilon = 10^{-4}$ to prevent the weight going to zero in the unlikely event of perfect recall.

We note that batches with examples 
from extremely rare classes
will have a much larger loss magnitude
than batches  with no such examples.
Hence, learning with small batches 
and large class imbalances
may lead to drastic changes in loss magnitude
across training iterations.

In multi-task learning, the total loss is calculated as the arithmetic mean over all tasks. 
Thus, a task with examples from rare classes may suffocate other tasks due to larger loss.
If we have many tasks that suffer from class imbalance, it is not unlikely that for any given batch, there will be one task that impedes the progress of other tasks.
This may prevent the model to learn any of the tasks, since they all intermittently hamper each other.
We address this problem by favoring a stable loss magnitude of individual tasks.
We achieve this by normalizing the loss with the
sum of the weights of individual examples.
If we reuse the weights $w^R_{ct}$ from equation \ref{eq:recall_loss},
then the loss for each individual task can be expressed as follows:

\begin{equation}
\mathrm{C E^{R}_{MT}}=\frac{-\sum_{n=1}^{N} w_{y_{n},t}^{R} \ln P_{n}^{y_{n}}}{\sum_{n=1}^{N} w_{y_{n},t}^{R}}.
\end{equation}

\subsection{Sequential enhancement}
\label{subsec:sequential}

The second stage of our recognition approach enhances 
local predictions 
by aggregating evidence
across a large temporal window.
We form temporal inputs
as sequences of T = 21 vectors
for all consecutive segments 
from (t-10) to (t+10). 
Instead of hand-crafted post-processing rules, we propose to classify prediction sequences 
with deep recurrent models \cite{yildirim18cbm}.
While the first stage of our approach predicted all attributes in a single forward pass, 
sequential enhancement involves 
per-attribute recurrent models.
Thus, these models can learn attribute-specific temporal behavior patterns from 
\ref{subsec:analysis_sequential}.
\begin{figure}[hb]
    \centering
    \includegraphics[width=\linewidth]
      {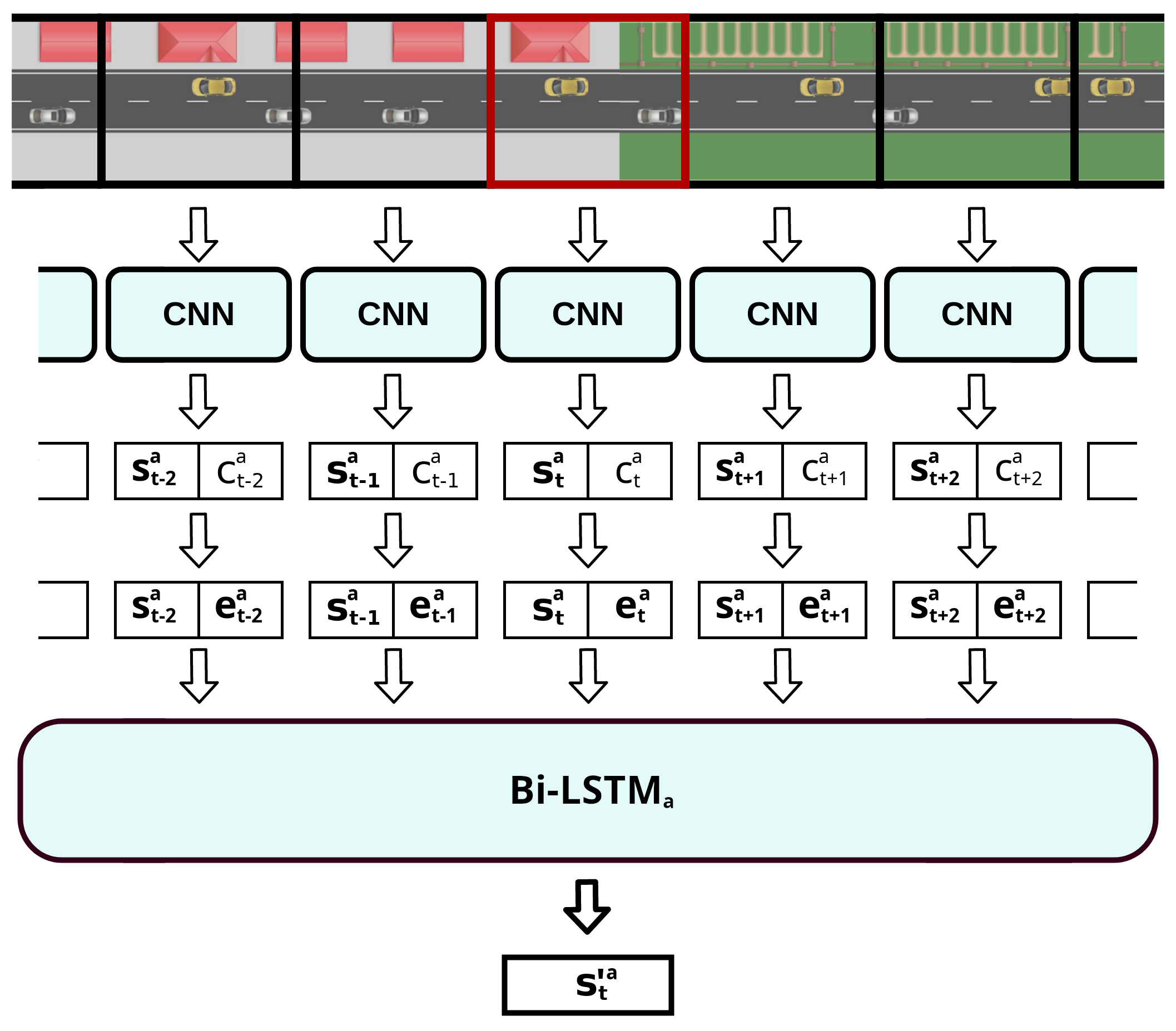}
    \caption{
  Sequential enhancement 
  corrects local predictions
  with per-attribute Bi-LSTM models.
  For each attribute $a$, the model $\mathrm{Bi\text{-}LSTM}_{a}$ outputs
  corrected logits $\mathbf{s'^a_t}$ in segment \textit{t}
  by observing \textit{T = 21} vectors
  that correspond to segments from \textit{(t-10)} to \textit{(t+10)}.
  Each of these vectors is a concatenation
  of the logits $\mathbf{s_i^a}$ and
  the jointly learned embedding $\mathbf{e_i^a}$
  of the the most probable class according to the local model.
}
    \label{fig:lstm}
\end{figure}

Our recurrent models consist of four layers 
with bidirectional long short-term memory cells (Bi-LSTM). 
Each layer processes the sequence in two directions using two separate unidirectional LSTM modules. 
The inputs are a concatenation 
of the local logits $\mathbf{s}^a_i$ 
and the embedding $\mathbf{e}_{c^a_i}$ 
of the winning class $c^{a}_{i} = \arg \max \mathbf{s}^a_i$. 
The matrices of per-attribute class embeddings 
are jointly learned with the entire recurrent model.
The dimensionality of the class embeddings 
is $\max(4, C)$ where $C$ denotes the number of classes. 
The dimension of hidden states in all layers is 128.
The last hidden state of a Bi-LSTM layer is the concatenation of the last hidden states produced by the two unidirectional LSTM modules.
The final representation of the input sequence is created by concatenating the last hidden state vectors of all layers, together with the hidden state of the middle element of the last layer.
This results in a vector of size 1280, which is then fed to a fully-connected softmax classifier.
The output is the predicted posterior distribution over classes $P(A_i=c^i_j | x_{t-T:t+T})$.
The whole model is trained with dynamically weighted cross-entropy as presented in \ref{subsec:loss}.

\section{Datasets}
\label{sec:datasets}

\subsection{iRAP-BH}
This paper introduces a novel corpus of georeferenced video 
that we have acquired along 194 public road sections (2300 km total) 
in Bosnia and Herzegovina for off-line road-safety assessment. 
All videos were recorded in 2704x2028 RGB format at 25 frames per second with a GoPro HERO4 Black camera.
An average road section comprises 1175 10-meter segments, while an average segment spans 18 frames.

Our corpus was annotated with all iRAP attributes by trained human annotators.
Even though the iRAP Star Rating Score 
requires 100-meter granularity, 
we have annotated iRAP-BH 
over 10-meter segments 
in order to provide better supervision 
for learning algorithms.
We split the dataset into 214,073 training, 5,813 validation, and 6,563 testing segments.
The 
three splits ensure that any two segments that belong to the same road section also belong to the same split.
This enables training sequential and multi-frame models without data leakage.
We represent each road segment with its middle frame resized to 384x288.
This results in a multi-task multi-class video recognition dataset with 226,449 images. 
Qualitative examples are provided 
in Figure \ref{fig:LSTMExamples}
and the appendix.

iRAP-BH is not related 
to any existing public dataset.
We will make it publicly available upon acceptance
in order to promote future research 
on road-safety and related tasks.

\subsection{Honda Scenes}
The Honda Scenes dataset \cite{narayanan19icra} contains 80 training and 20 evaluation videos.
Each frame of each video is annotated 
for the following four 
traffic scene classification problems:
\attribute{Road place}, 
\attribute{Road environment}, 
\attribute{Road surface}, 
\attribute{Weather}.

Images for the Road place and Road environment problems have been obtained 
by subsampling at 3Hz. 
This results in 760,000 training 
and 160,000 evaluation frames.
We evaluate our methods by treating consecutive 
frames as consecutive road segments. 
Images for the Road surface and Weather problems  
have been sampled from Honda Scenes and BDD100k. 
The Road surface dataset consists of 
2,676 Honda and 7,463 BDD100k images.
The training and evaluation splits contain 
9,150 and 898 images.
The Weather dataset is a subset of BDD100k 
with 11,781 training and 1,255 test images.
The following paragraphs briefly describe 
the four problems.

\subsubsection{Road place}
This is the only problem that considers multiple multi-class classification tasks. 
Each of those tasks has fine-grained temporal labels with classes such as Approaching (A), Entering (E), and Passing (P) that depend on the relative position of the car to the place of interest in a given frame.
The tasks are:
\attribute{Construction zone}, \attribute{Intersection (3 way)}, \attribute{Intersection (4 way)},
\attribute{Intersection (5 way \& more)},
\attribute{Overhead bridge}, \attribute{Rail crossing},
\attribute{Merge - Gore On Left}, \attribute{Merge - Gore On Right}, 
\attribute{Branch - Gore On Left}, \attribute{Branch - Gore On Right}, 
\attribute{Background}.
These tasks are related to iRAP-BH 
since they involve sequential recognition 
of road infrastructure elements.

\subsubsection{Road environment}
This problem involves recognition of the following classes: \attribute{Local}, \attribute{Highway}, \attribute{Ramp}, \attribute{Urban}.
The problem does not involve temporal labels, 
but it consists entirely of frames 
from the Honda Scenes dataset.
That means we can still use 
our multi-frame and sequential models.

\subsubsection{Road surface}
In this multi-class problem, each image is annotated with one of three road surface classes: \attribute{Wet}, \attribute{Dry}, and \attribute{Snow}.
We evaluate 
only the single-frame version of our model 
since the dataset includes only  
non-sequential images.
The \attribute{Road surface} classes are fairly balanced, so weighted losses do not improve performance.

\subsubsection{Weather}
This multi-class problem involves image classification into four classes: \attribute{Clear}, \attribute{Overcast}, \attribute{Rainy}, and \attribute{Snowy}. 
We evaluate only our unbalanced single-frame model 
for the same reasons as in the case of \attribute{Road surface}. 

\subsection{FM3m}
The third iteration of the Fleet Management Dataset (FM3) \cite{sikiric20tits} consists of 11,448 images of traffic scenes from Croatian roads. 
The main subset of the dataset (FM3m) consists of 6,413 images. 
The training, validation and test splits contain 1,607, 1,600, and 3,206 images, respectively.
Each image is labeled with one binary label (true/false) for each classification attribute. 
There are 8 classification binary attributes: \attribute{highway}, \attribute{road}, \attribute{tunnel}, \attribute{exit}, \attribute{settlement}, \attribute{overpass}, \attribute{booth}, \attribute{traffic}.

The frames were assigned to training, validation, and test splits in a uniform random fashion.
This means that, in general, consecutive frames were assigned to different splits. 
This prevents us to take 
multiple segments on input, 
so we evaluate only single-frame models 
in these experiments.

\subsection{BDD100k}
The Berkeley Deep Drive (BDD100k) dataset \cite{yu20cvpr} 
has been designed for heterogeneous 
multi-task visual recognition 
in road driving scenes. 
It contains 100,000 40-second video clips 
from a wide range of environments,
including urban and rural areas, 
different weather conditions and times of day.
From each sequence, a single keyframe 
was annotated 
with object bounding boxes, drivable areas, 
lane markings and full-frame panoptic labels.  
There are 80,000 training, 10,000 validation, 
and 20,000 test images, respectively.
Additionally, there are three image-wide tasks: 
\attribute{Scene}, \attribute{Weather}, and \attribute{Time of day}.
The \attribute{Scene} classes include \attribute{Tunnel}, \attribute{Residential}, \attribute{Parking Lot}, \attribute{City Street}, \attribute{Countryside}, \attribute{Gas Station}, and \attribute{Highway}.
Our experiments focus on the \attribute{Scene} task since the other two tasks 
have not been addressed 
by relevant previous work. 
Moreover, there is an intersection 
between the \attribute{Weather} task 
and the homonymous task of Honda Scenes.

BDD100k does not allow sequential processing
since each image comes 
from a different video sequence.
Thus, our experiments involve single-frame models, as in FM3m, and the Road surface and Weather problems of Honda Scenes.

\section{Experimental Results and Analysis}
\label{sec:experiments}

We evaluate visual recognition of road-safety attributes
with different variants of our approach on the iRAP-BH dataset. 
Moreover, we compare our approach with previous work on three related datasets: Honda Scenes, FM3m, and BDD100k.
Though not annotated for iRAP attributes, these three datasets 
involve classification of road and traffic-related classes, some of which are quite similar to certain iRAP attributes.

\subsection{Evaluation metrics}
\label{subsec:metrics}

We evaluate our approaches on iRAP-BH and Honda Scenes according to the mean macro-averaged F1 score \cite{f1,opitz19corr}.
This is a suitable metric due to our 
multi-task multi-class setup 
and class imbalance.
Creators of Honda Scenes also use macro-F1 in their experiments.
We evaluate FM3m performance according to the 
mean average precision (mAP) across all tasks.
This metric is suitable since all tasks involve binary classification \cite{sikiric20tits}.
We evaluate BDD100k classification performance according to 
accuracy \cite{ni22tim, saffari23tetci}. 
We present all performance metrics as percentage points (pp).

\subsection{iRAP-BH}

We 
augment input images through color jittering by 
varying brightness, contrast, saturation, and hue in relative ratios of 0.6, 0.3, 0.2, and 0.02. 
We do not employ horizontal flipping as it would 
disturb the detection of 
attributes that 
are specific to right-hand traffic scenarios (e.g. \attribute{Roadside severity - passenger side}). 
We also opt not to apply random cropping as it omits visual information from the peripheral parts of input images 
since this information is 
critical for identifying roadside attributes such as \attribute{Street lighting}. 
Both stages of our method are trained with the Adam optimizer.
For 
local recognition, 
the learning rate is set to 1e-5, weight decay to 1e-3 and batch size to 12. 
We train for 15 epochs according to a multiplicative learning rate scheduler with the 
annealing factor of 0.88 per epoch. 
For sequential enhancement we set  
the learning rate to 5e-4, weight decay to 1e-4, and train for 10 epochs with batch size 32.
We have systematically explored multiple values for each hyperparameter involved in our model architecture and the optimization process. 
The optimal hyperparameter values were selected by conducting two separate grid searches on the validation split of iRAP-BH. 
The two grid searches optimized hyperparameters
of the two stages of our method. 
The optimal values are consistently applied in all subsequent experiments, across all datasets. 
A complete list of 
hyperparameters and their values is provided in the appendix.

Table \ref{bih_mean_comparison} explores 
the impact of sequential post-processing and loss weighting on the overall performance. 
We observe that semantic segmentation pre-training 
contributes 1.2 percentage points (pp) mF1.
Inverse-frequency weighting delivers additional 1.5 \,pp. 
Our multi-task weighting improves upon that by 1.3\,pp. 
Finally, sequential post-processing further increases the performance by about 5.1\,pp.
\begin{table}[b]
\centering
\caption{
  Impact of 
  our contributions 
  on overall Macro-F1 performance on iRAP-BH test.
    IFW - inverse frequency weighting;
    R - dynamic recall weighting;  
    SE - sequential enhancement.
}
\label{bih_mean_comparison}
\begin{tabular}{|l|c|c|}
\hline
Model                            & Pre-training & \multicolumn{1}{l|}{Macro-F1} \\ \hline
$\mathrm{ConvCE}$              & ImageNet     & 53.79                         \\
$\mathrm{ConvCE}$              & Vistas       & 54.96                         \\
$\mathrm{ConvCE}^{\text{IFW}}_{\text{MT}}$   & Vistas       & 56.43                         \\
$\mathrm{ConvCE}^{\text{R}}_{\text{MT}}$     & Vistas       & 57.77                         \\
$\mathrm{ConvCE}\text{-}\mathrm{SE}$          & Vistas       & 59.68                         \\
$\mathrm{ConvCE}^{\text{R}}_{\text{MT}}\text{-}\mathrm{SE}$ & Vistas       & \textbf{62.86}                \\ \hline
\end{tabular}
\end{table}

Table \ref{bih_per_attribute} 
shows the impact of sequential enhancement 
and dynamic loss weighting on each attribute.
The attributes which get the largest relative improvement from loss weighting are \attribute{Pedestrian crossing - inspected road}, \attribute{Median Type}, \attribute{Pedestrian observed flow along the road passenger-side}, \attribute{Roadside severity - driver-side object}, and \attribute{Bicycle facility}.
Relative improvements compared to standard cross-entropy for those attributes range from 19\% to 26.5\%.
All of these attributes suffer from class imbalance.

\begin{table}[h]
\centering
\caption{Per-attribute mean macro-F1 performance of two ablated models and our best model on iRAP-BH test.
}
\label{bih_per_attribute}
\resizebox{\columnwidth}{!}{%
\begin{tabular}{|l|c|c|c|c|}
\hline
Attribute                                     & CNN $\mathrm{CE}$ & +LSTM $\mathrm{CE}$ & +LSTM $\mathrm{CE}^{\text{R}}_{\text{MT}}$  \\ \hline
\hline
Area type                                              & 90.49 & 90.93 & 94.56     \\ \hline
Bicycle facility                                       & 53.72 & 100.00 & 100.00    \\ \hline
Bicycle observed flow                                  & 35.77 & 35.82 & 34.83     \\ \hline
Carriageway label                                      & 93.88 & 94.20 & 97.54     \\ \hline
Curvature                                              & 55.22 & 59.55 & 64.77     \\ \hline
Delineation                                            & 98.12 & 99.17 & 98.94     \\ \hline
Grade                                                  & 53.60 & 53.63 & 51.85     \\ \hline
Intersection channelisation                            & 61.55 & 63.41 & 62.74     \\ \hline
Intersection quality                                   & 47.83 & 51.14 & 52.02     \\ \hline
Intersection type                                      & 27.39 & 33.51 & 34.97     \\ \hline
Land use - driver                                      & 62.35 & 64.46 & 63.85     \\ \hline
Land use - passenger                                   & 62.81 & 67.08 & 69.60     \\ \hline
Lane width                                             & 64.99 & 71.21 & 75.60     \\ \hline
Median Type                                            & 40.76 & 44.45 & 55.62     \\ \hline
Motorcycle observed flow                               & 26.04 & 33.23 & 36.45     \\ \hline
Number of lanes                                        & 71.09 & 84.32 & 98.49     \\ \hline
Paved shoulder - driver                                & 63.54 & 64.71 & 65.02     \\ \hline
Paved shoulder - passenger                             & 46.43 & 48.52 & 54.01     \\ \hline
Ped. crossing - inspected rd.                          & 35.61 & 35.73 & 45.03     \\ \hline
Ped. crossing - side rd.                               & 39.12 & 43.78 & 46.58     \\ \hline
Ped. crossing quality                                  & 48.11 & 53.11 & 59.21     \\ \hline
Ped. obs. flow, across                                 & 26.22 & 33.54 & 44.31     \\ \hline
Ped. obs. flow, driver                                 & 24.73 & 27.61 & 29.37     \\ \hline
Ped. obs. flow, pass.                                  & 26.96 & 28.74 & 35.62     \\ \hline
Property access points                                 & 56.64 & 57.28 & 59.71     \\ \hline
Quality of curve                                       & 65.46 & 65.99 & 71.24     \\ \hline
Road condition                                         & 72.76 & 73.32 & 77.70     \\ \hline
Roadside severity, driver dist.                        & 56.94 & 57.05 & 57.86     \\ \hline
Roadside severity, driver obj.                         & 35.25 & 35.26 & 43.31     \\ \hline
Roadside severity, pass. dist.                         & 54.90 & 56.87 & 59.97     \\ \hline
Roadside severity, pass. obj.                          & 46.94 & 50.51 & 51.06     \\ \hline
Roadworks                                              & 70.61 & 74.29 & 78.80     \\ \hline
Sc. zone crossing supervisor                           & 62.84 & 65.60 & 66.51     \\ \hline
Sc. zone warning                                       & 62.57 & 66.90 & 68.03     \\ \hline
Service road                                           & 55.28 & 59.44 & 62.26     \\ \hline
Sidewalk - driver-side                                 & 43.63 & 46.08 & 43.72     \\ \hline
Sidewalk - passenger-side                              & 44.31 & 45.83 & 50.19     \\ \hline
Sight distance                                         & 70.00 & 70.85 & 75.67     \\ \hline
Skid resistance / grip                                 & 38.77 & 44.52 & 50.90     \\ \hline
Speed management                                       & 61.57 & 100.00 & 100.00    \\ \hline
Street lighting                                        & 90.58 & 91.38 & 91.53     \\ \hline
Upgrade cost                                           & 61.48 & 62.45 & 63.28     \\ \hline
Vehicle parking                                        & 56.29 & 60.57 & 60.28     \\ \hline
\hline
Mean                                          & 54.96 & 59.68 & 62.86 \\ \hline
\end{tabular}%
}
\end{table}

\begin{figure*}[t]
  \includegraphics[width=\textwidth]{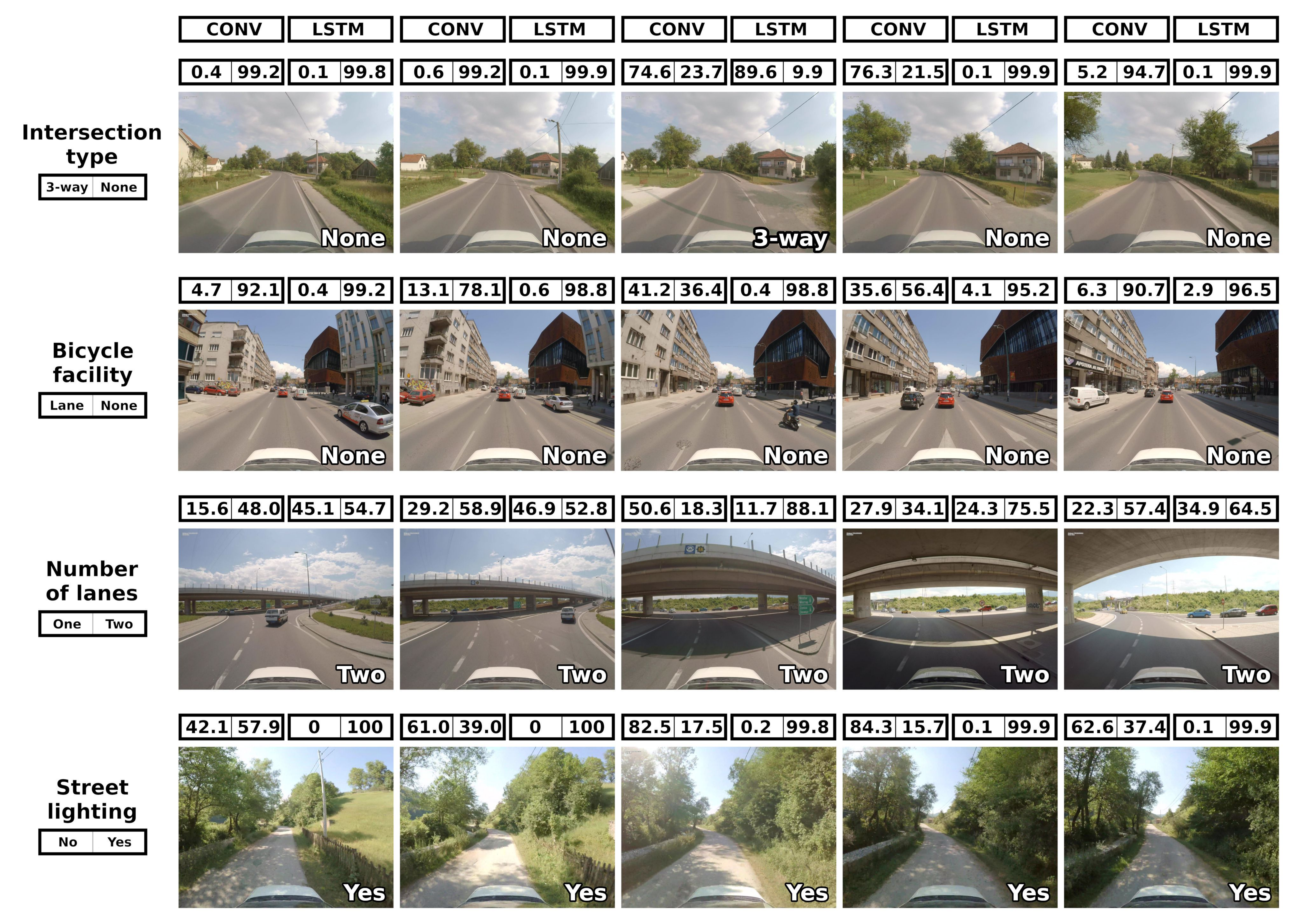}
  \caption{
  Four iRAP-BH examples where sequential enhancement (LSTM)
  succeeds to correct our local visual predictions (CNN).
  For each of the five consecutive segments (columns),
  we display categorical predictions by both models (top) and the ground truth label (bottom right).
  Row 1 involves a single-peak attribute - \attribute{Intersection type}.
  We observe that the local model incorrectly assigns
  a positive class (\attribute{3-way intersection}) in column 4.
  Rows 2 and 3 involve smooth attributes -
  \attribute{Bicycle facility} and \attribute{Number of lanes}.
  We observe that the local model
  mistakes a motorcyclist near a tram rail
  for a dedicated bicycle lane in column 3
  and fails to predict the correct number of lanes again in column 3.
  Row 4 involves the \attribute{Street lighting} attribute.
  We observe that the upcoming lighting poles are obscured
  by the road curvature and overgrown roadside bushes.
  Consequently, the local model fails in columns 2-4.
  In all cases, the sequential model succeeds
  to correct the mistakes by leveraging a larger temporal context.
}
 \label{fig:LSTMExamples}
\end{figure*}

Single-peak attributes which benefit most 
from sequential enhancement are 
\attribute{Speed management / traffic calming}
and \attribute{Intersection type}, 
with relative improvements of 44.7\% and 22.3\%.
Smooth attributes with the largest 
relative improvements are 
\attribute{Bicycle facility}, 
\attribute{Number of lanes}, and 
\attribute{Skid resistance / grip}.
Figure \ref{fig:LSTMExamples} shows four examples 
of successful sequential enhancement.
In the case of \attribute{Intersection type}, 
the correction accommodated 
the single-peak annotation convention.
For smooth attributes 
\attribute{Bicycle facility} and 
\attribute{Number of lanes}, 
the sequential model corrected 
the spurious class transition 
made by the local model
by considering a larger context.
The last example is \attribute{Street lighting},
which should be annotated continuously 
through all segments 
between two nearby lighting poles.
In this particular example, 
the upcoming lighting poles 
are obscured by the road curvature 
and overgrown roadside bushes.
Consequently, the local model classifies 
the in-between segments incorrectly. 
The sequential model 
corrects these mistakes
by leveraging a larger context window.  

Table \ref{table:pretraining_ablation} 
validates several pre-training strategies
for the backbone of our model.
We observe that dense semantic pre-training
outperforms classification pre-training.
This suggests that detailed 
spatial understanding of visual concepts 
benefits the recognition 
of road-safety attributes.

\begin{table}[h]
\centering
\caption{Impact of different pre-training strategies on overall Macro-F1 performance on iRAP-BH test.}
\label{table:pretraining_ablation}
\begin{tabular}{|l|l|c|}
\hline
Dataset & Task         & Macro-F1 \\ \hline
ImageNet-1k      & Classification        & 61.26                  \\
BDD100k          & Classification        & 61.19                  \\
Honda Scenes     & Classification        & 61.35                  \\
Vistas           & Semantic Segmentation & \textbf{62.86}         \\ \hline
\end{tabular}%
\end{table}

\subsection{Honda Scenes}

This subsection compares 
our method with 
prominent previous work 
on Honda Scenes.
We include several methods 
from the original paper
\cite{narayanan19icra},
Context MTL \cite{lee20iv}, 
MTAN \cite{liu19cvpr} 
and two our ablations that 
show the impact of our contributions.

\subsubsection{Road place}

Table \ref{honda_road_place_overall} 
evaluates the overall performance.
We denote the original 
sequential baseline  
as Honda BiLSTM 
and their 
two-stage sequential approach 
as Honda Event
\cite{narayanan19icra}.

\begin{table}[htb]
\centering
\caption{Macro-F1 performance on Honda Scenes - Road-place.}

\label{honda_road_place_overall}
\begin{tabular}{|l|c|cc|}
\hline
\multicolumn{1}{|c|}{\multirow{2}{*}{Model}} & \multirow{2}{*}{BB} & \multicolumn{2}{c|}{Road place} \\ \cline{3-4} 
\multicolumn{1}{|c|}{}                                &                              & Mean                & Mean w/o Background         \\ \hline
Honda BiLSTM \cite{narayanan19icra}                   & rn50                         & 27.56               & 25.23              \\
Honda Event  \cite{narayanan19icra}                   & rn50                         & 28.36               & 25.91              \\
Context MTL  \cite{lee20iv}                           & rn50                         & -                   & 27.92              \\
MTAN  \cite{liu19cvpr}                           & wrn28                        & 29.14          & 26.73    \\
$\mathrm{ConvCE}$ (ours)                            & rn18                         & 34.11               & 31.96              \\
$\mathrm{ConvCE}^{\text{R}}_{\text{MT}}$ (ours)                   & rn18                         & 37.00               & 34.92              \\
$\mathrm{ConvCE}^{\text{R}}_{\text{MT}}\text{-}\mathrm{SE}$ (ours)               & rn18                         & \textbf{40.93}      & \textbf{39.00}     \\ \hline
\end{tabular}
\end{table}

Table \ref{honda_road_place_subtasks} 
focuses on individual tasks.
In both tables, our baseline 
(multi-frame model, standard loss, 
 no sequential enhancement) 
outperforms all previous approaches 
in spite of a weaker backbone. 
This makes the improvements in Tables 
\ref{bih_mean_comparison} 
--
\ref{table:pretraining_ablation} 
even more convincing since now 
we know that they start 
from a very strong baseline. 
Multi-task dynamic loss weighting 
increases our performance by 2.9 pp.
This 
improvement is due to class imbalance \cite{narayanan19icra}.
We observe the greatest 
relative improvement 
on \attribute{Intersection (5-way or more)}, \attribute{Railway}, \attribute{Left Merge}, and \attribute{Left Branch} subtasks.
Sequential enhancement brings a further improvement of 3.9 pp.
\begin{table*}[t]
\centering
\caption{Experimental evaluation on all Road-Place tasks of Honda Scenes (macro-F1, percentage points).
  Legend: BB - backbone;
  B-\attribute{Background}, A-\attribute{Approaching},
  E-\attribute{Entering}, P-\attribute{Passing}.}
\label{honda_road_place_subtasks}
\resizebox{\textwidth}{!}{
\begin{tabular}{|l|c|c|cccc|cccc|cccc|ccc|ccc|}
\hline
\multicolumn{1}{|c|}{\multirow{2}{*}{Model}} & \multirow{2}{*}{BB} & B            & \multicolumn{4}{c|}{Intersection 5-way} & \multicolumn{4}{c|}{Railway Crossing}     & \multicolumn{4}{c|}{Construction}         & \multicolumn{3}{c|}{Left Merge}  & \multicolumn{3}{c|}{Right Merge}  \\
\multicolumn{1}{|c|}{}                                &                              & -                     & A        & E        & P        & {\ul Mean}        & A        & E        & P        & {\ul Mean}        & A        & E        & P        & {\ul Mean}        & A          & P        & {\ul Mean}        & A          & P         & {\ul Mean}        \\ \hline
Honda BiLSTM  \cite{narayanan19icra}                  & rn50                         & 88                    & 0        & 0        & 9        & 3                 & 24       & 14       & 46       & 28                & 2        & 5        & 29       & 12                & 9          & 28       & 19                & 16         & 23        & 20                \\
Honda Event  \cite{narayanan19icra}                   & rn50                         & \textbf{92}           & 0        & 0        & 0        & 0                 & 23       & 47       & 46       & 39                & 2        & 6        & 38       & 15                & 5.6        & 8        & 7                 & 13         & 16        & 15                \\
Context MTL  \cite{lee20iv}                           & rn50                         & -                     & 0        & 6        & 0        & 2                 & 1        & 35       & 52       & 32                & 0        & 4        & 38       & 14                & 4          & 6        & 5                 & 26         & 18        & 22                \\
MTAN  \cite{liu19cvpr}                           & wrn28                        & 92                     & 1        & 2        & 5        & 3                 & 19        & 27        & 42        & 29                 & 3        & 9        & 24        & 12                 & 11          & 17        & 14                 & 19          & 12         & 16                 \\
$\mathrm{ConvCE}$ (ours)                            & rn18                         & 90                    & 19       & 0        & 5        & 8                 & 13       & 49       & 52       & 38                & 2        & 11       & 56       & 23                & 22         & 29       & 26                & 29         & 33        & 31                \\
$\mathrm{ConvCE}^{\text{R}}_{{\text{MT}}}$ (ours)    & rn18                         & 91                    & 27       & 0        & 10       & 12                & 15       & 56       & 59       & 43                & 3        & 12       & 63       & 26                & 27         & 36       & 32                & 31         & 35        & 33                \\
$\mathrm{ConvCE}^{\text{R}}_{{\text{MT}}}\text{-}\mathrm{SE}$ (ours)               & rn18                         & 91                    & 29       & 0        & 9        & \textbf{13}       & 28       & 53       & 71       & \textbf{51}       & 11       & 22       & 64       & \textbf{32}       & 29         & 43       & \textbf{36}       & 34         & 45        & \textbf{40}       \\ \hline
\multicolumn{1}{|c|}{\multirow{2}{*}{Model}} & \multirow{2}{*}{BB} & \multicolumn{1}{l|}{} & \multicolumn{4}{c|}{Overhead Bridge}      & \multicolumn{4}{c|}{Intersection 3-way} & \multicolumn{4}{c|}{Intersection 4-way} & \multicolumn{3}{c|}{Left Branch} & \multicolumn{3}{c|}{Right Branch} \\
\multicolumn{1}{|c|}{}                                &                              & \multicolumn{1}{l|}{} & A        & E        & P        & {\ul Mean}        & A        & E        & P        & {\ul Mean}        & A        & E        & P        & {\ul Mean}        & A          & P        & {\ul Mean}        & A          & P         & {\ul Mean}        \\ \hline
Honda BiLSTM   \cite{narayanan19icra}                 & rn50                         &                       & 23       & 55       & 53       & 44                & 3        & 28       & 27       & 19                & 14       & 68       & 66       & 49                & 36         & 22       & 29                & 28         & 28        & 28                \\
Honda Event   \cite{narayanan19icra}                  & rn50                         &                       & 42       & 58       & 59       & 53                & 8        & 16       & 23       & 16                & 31       & 7        & 67       & 56                & 30         & 19       & 25                & 24         & 22        & 23                \\
Context MTL \cite{lee20iv}                            & rn50                         &                       & 47       & 59       & 60       & \textbf{55}       & 11       & 38       & 28       & 26                & 14       & 78       & 79       & 57                & 33         & 19       & 26                & 34         & 27        & 31                \\
MTAN  \cite{liu19cvpr}                           & wrn28                        &                       & 31        & 49        & 57        & 46                 & 15        & 35        & 31        & 27                 & 25        & 75        & 73        & 58                 & 38          & 26        & 32                 & 30          & 19         & 25                 \\
$\mathrm{ConvCE}$ (ours)                            & rn18                         &                       & 31       & 57       & 60       & 50                & 10       & 33       & 32       & 25                & 25       & 77       & 66       & 56                & 29         & 18       & 24                & 35         & 38        & 37                \\
$\mathrm{ConvCE}^{\text{R}}_{{\text{MT}}}$ (ours)                   & rn18                         &                       & 33       & 58       & 61       & 51                & 10       & 34       & 34       & 26                & 26       & 77       & 68       & 57                & 31         & 23       & 27                & 37         & 42        & 40                \\
$\mathrm{ConvCE}^{\text{R}}_{{\text{MT}}}\text{-}\mathrm{SE}$ (ours)               & rn18                         &                       & 37       & 59       & 64       & 53                & 14       & 32       & 38       & \textbf{28}       & 30       & 78       & 73       & \textbf{60}       & 42         & 24       & \textbf{33}       & 41         & 44        & \textbf{43}       \\ \hline
\end{tabular}
}
\end{table*}

\subsubsection{Road environment}
Table \ref{honda_other_tasks_per_class} 
compares our multi-frame model 
with Context MTL, MTAN, and the two 
original frame-based approaches
\cite{narayanan19icra}. 
These approaches involve a ResNet-50 backbone 
pre-trained on Places365 and leverage
the DeepLabV2 semantic segmentation model 
to either mask out traffic participants 
(Honda Frame - Mask) or 
enrich the input image 
with its segmentation map 
(Honda Frame - SemSeg).
Our model prevails on most classes.
The largest improvement occurs 
on the most challenging class (\attribute{Ramp}).
That is the least frequent class 
in the dataset, and the class that benefits the most
from multi-task loss weighting.
Sequential enhancement brings 
a further improvement of 1.2 pp.

\begin{table}[H]
\centering
\caption{Macro-F1 performance on three problems  
of Honda Scenes.}
\label{honda_other_tasks_per_class}
\resizebox{\columnwidth}{!}{%
\begin{tabular}{|lccccccc|}
\hline
\multicolumn{8}{|c|}{Road environment}                                                                                                                                                            \\ \hline
\multicolumn{1}{|c|}{Model}                        & \multicolumn{1}{c|}{BB} & Local         & Highway       & Ramp           & Urban         & \multicolumn{2}{c|}{{\ul Mean}}          \\ \hline
\multicolumn{1}{|l|}{Honda Frame - Mask \cite{narayanan19icra}}    & \multicolumn{1}{c|}{rn50}        & 33.0          & 91.0          & 20.0           & 83.0          & \multicolumn{2}{c|}{56.8}                \\
\multicolumn{1}{|l|}{Honda Frame - SemSeg \cite{narayanan19icra}}    & \multicolumn{1}{c|}{rn50}        & 34.0          & 89.0          & 13.0           & 81.0          & \multicolumn{2}{c|}{54.3}                \\
\multicolumn{1}{|l|}{Context MTL \cite{lee20iv}}            & \multicolumn{1}{c|}{rn50}        & 36.0 & 92.0          & 21.0           & 81.0          & \multicolumn{2}{c|}{57.5}                \\
\multicolumn{1}{|l|}{MTAN \cite{liu19cvpr}}            & \multicolumn{1}{c|}{wrn28}        & \textbf{37.2} & 91.1          & 19.7           & 80.3          & \multicolumn{2}{c|}{57.1}                \\
\multicolumn{1}{|l|}{$\mathrm{ConvCE}$ (ours)}     & \multicolumn{1}{c|}{rn18}        & 29.1          & 91.2          & 36.3           & 82.9          & \multicolumn{2}{c|}{59.9}                \\
\multicolumn{1}{|l|}{$\mathrm{ConvCE}^\text{R}_{\text{MT}}$ (ours)}     & \multicolumn{1}{c|}{rn18}        & 30.8          & 92.3          & 42.6           & 83.7          & \multicolumn{2}{c|}{62.4}                \\
\multicolumn{1}{|l|}{$\mathrm{ConvCE}^\text{R}_{\text{MT}}\text{-}\mathrm{SE}$ (ours)} & \multicolumn{1}{c|}{rn18}        & 32.8          & \textbf{93.1} & \textbf{44.1}  & \textbf{84.2} & \multicolumn{2}{c|}{\textbf{63.6}}       \\ \hline
\multicolumn{8}{|c|}{Road surface}                                                                                                                                                                \\ \hline
\multicolumn{1}{|c|}{Model}                        & \multicolumn{1}{c|}{BB} & Dry           & Wet           & Snow           & {\ul Mean}    & \multicolumn{2}{c|}{{\ul Mean w/o Snow}} \\ \hline
\multicolumn{1}{|l|}{Honda Frame - Mask \cite{narayanan19icra}}    & \multicolumn{1}{c|}{rn50}        & 93.0          & 92.0          & 99.7           & 94.9          & \multicolumn{2}{c|}{92.5}                \\
\multicolumn{1}{|l|}{Honda Frame - SemSeg \cite{narayanan19icra}}    & \multicolumn{1}{c|}{rn50}        & 92.2          & 92.5          & 99.0           & 94.6          & \multicolumn{2}{c|}{92.4}                \\
\multicolumn{1}{|l|}{Context MTL \cite{lee20iv}}            & \multicolumn{1}{c|}{rn50}        & 93.0          & 92.0          & -              & -             & \multicolumn{2}{c|}{92.5}                \\
\multicolumn{1}{|l|}{MTAN \cite{liu19cvpr}}            & \multicolumn{1}{c|}{wrn28}        & 94.2 & 94.1          & 98.9           & 95.7          & \multicolumn{2}{c|}{94.2}                \\
\multicolumn{1}{|l|}{Conv single (ours)}                    & \multicolumn{1}{c|}{rn18}        & \textbf{98.5} & \textbf{98.5} & \textbf{100.0} & \textbf{99.0} & \multicolumn{2}{c|}{\textbf{98.5}}       \\ \hline
\multicolumn{8}{|c|}{Weather}                                                                                                                                                                     \\ \hline
\multicolumn{1}{|c|}{Model}                        & \multicolumn{1}{c|}{BB}          & Clear         & Overcast      & Rain           & Snow          & {\ul Mean}       & {\ul Mean w/o Snow}   \\ \hline
\multicolumn{1}{|l|}{Honda Frame - Mask \cite{narayanan19icra}}    & \multicolumn{1}{c|}{rn50}        & 92.0          & 83.0          & 96.0           & 94.0          & 91.3             & 90.3                  \\
\multicolumn{1}{|l|}{Honda Frame - SemSeg \cite{narayanan19icra}}    & \multicolumn{1}{c|}{rn50}        & 91.6          & 83.4          & 96.0           & 93.9          &      91.2        & 90.3                  \\
\multicolumn{1}{|l|}{Context MTL \cite{lee20iv}}            & \multicolumn{1}{c|}{rn50}        & \textbf{93.2} & 84.0          & \textbf{97.0}  & -             & -                & 91.4                  \\
\multicolumn{1}{|l|}{MTAN \cite{liu19cvpr}}            & \multicolumn{1}{c|}{wrn28}        & 90.4         & 85.9           & 91.2          & 92.5   &  90.0 & 89.2                 \\
\multicolumn{1}{|l|}{Conv single (ours)}                    & \multicolumn{1}{c|}{rn18}        & 91.0          & \textbf{90.9} & 95.0           & \textbf{95.3} & \textbf{93.0}    & \textbf{92.3}         \\ \hline
\end{tabular}%
}
\end{table}

\subsubsection{Road surface}

Table \ref{honda_other_tasks_per_class} shows that our single-frame model outperforms 
previous approaches.
Loss weighting does not improve performance 
since the classes are fairly balanced.
We could not use our multi-frame 
and sequential enhancements since 
the task only allows single-frame prediction.

Table \ref{honda_surface_weather_ablation} shows that pre-training the ResNet-18 
for semantic segmentation on Vistas 
works better than pre-training the ResNet-50 
for classification on ImageNet-1k.

\subsubsection{Weather}

Table \ref{honda_other_tasks_per_class} 
shows that our single-frame model 
outperforms previous approaches 
on classes \attribute{Overcast}, 
\attribute{Snow} and overall, 
while underperforming on 
\attribute{Clear} and \attribute{Rain}. 
As in the Road Surface task, 
loss weighting does not contribute
since the classes are balanced, 
while sequential approaches are not applicable 
to the single-frame prediction task.

\begin{table}[h]
\centering
\caption{Ablation of segmentation pre-training (Macro-F1) on Road surface and Weather problems of the Honda Scenes dataset.}
\label{honda_surface_weather_ablation}
\begin{tabular}{|lcccccc|}
\hline
\multicolumn{7}{|c|}{Road surface}                                                                                                 \\ \hline
\multicolumn{1}{|c|}{Pre-training} & \multicolumn{1}{c|}{BB} & Dry   & Wet      & Snow  & \multicolumn{2}{c|}{{\ul Mean}} \\ \hline
\multicolumn{1}{|l|}{IN-1k (ours)}          & \multicolumn{1}{c|}{rn18}        & 97.2  & 96.8     & 99.7  & \multicolumn{2}{c|}{97.9}       \\
\multicolumn{1}{|l|}{IN-1k (ours)}          & \multicolumn{1}{c|}{rn50}        & 97.7  & 97.3     & 99.7  & \multicolumn{2}{c|}{98.2}       \\
\multicolumn{1}{|l|}{Vistas (ours)}         & \multicolumn{1}{c|}{rn18}        & \textbf{98.5}  & \textbf{98.5}     & \textbf{100.0} & \multicolumn{2}{c|}{\textbf{99.0}}       \\ \hline
\multicolumn{7}{|c|}{Weather}                                                                                                      \\ \hline
\multicolumn{1}{|c|}{Pre-training} & \multicolumn{1}{c|}{BB} & Clear & Overcast & Rain  & Snow        & {\ul Mean}        \\ \hline
\multicolumn{1}{|l|}{IN-1k (ours)}          & \multicolumn{1}{c|}{rn18}        & 89.5  & 86.8     & 94.0  & 93.0        & 90.8              \\
\multicolumn{1}{|l|}{IN-1k (ours)}          & \multicolumn{1}{c|}{rn50}        & 90.1  & 87.6     & 94.3  & 94.0        & 91.5              \\
\multicolumn{1}{|l|}{Vistas (ours)}         & \multicolumn{1}{c|}{rn18}        & \textbf{91.0}  & \textbf{90.9}     & \textbf{95.0}  & \textbf{95.3}        & \textbf{93.0}              \\ \hline
\end{tabular}
\end{table}

\subsection{FM3m}

We compare our method with two of the best approaches provided by the authors of the FM3m dataset.
They train SVM classifiers with RBF kernels on image descriptors obtained by ResNet-50 and DenseNet-121 backbones pre-trained on ImageNet-1k.
It should be noted that the 
two approaches do not fine-tune 
their feature extractors on FM3m,
while our approach has a weaker backbone.
We also compare against other concurrent approaches for traffic scene classification, namely the two variants of the Honda Frame \cite{narayanan19icra} frame-based model and the multi-task attention network (MTAN) \cite{liu19cvpr}.
Table \ref{fm3m_vs_their} shows that our single-frame models 
perform competitively, 
and that our multi-task loss-weighting improves performance on nearly all classes.

\begin{table}[h]
\centering
\caption{AP performance on the FM3m dataset: H-\attribute{Highway}, R-\attribute{Road}, Tu-\attribute{Tunnel}, E-\attribute{Exit}, S-\attribute{Settlement}, O-\attribute{Overpass}, B-\attribute{Booth}, Tr-\attribute{Traffic}.}
\label{fm3m_vs_their}
\resizebox{\columnwidth}{!}{%
\begin{tabular}{|l|ccccccccc|}
\hline
\multicolumn{1}{|c|}{Model} & H              & R              & Tu             & E             & S             & O             & B             & Tr            & \uline{Mean}  \\ \hline
RN50-SVM \cite{sikiric20tits}        & 99.8           & 91.1           & \textbf{100.0} & 97.7          & 98.3          & 97.2          & 98.8          & 86.8          & 96.2          \\
DN121-SVM \cite{sikiric20tits}       & 93.7           & \textbf{100.0} & 98.0           & 98.3          & 97.9          & 98.8          & 87.9          & \textbf{96.8} & 96.4          \\
Honda Frame - Mask \cite{narayanan19icra}       & 99.5           & 91.0      & 99.2           & 93.1          & 97.1          & 96.5          & 97.9          & 80.4 & 94.4          \\
Honda Frame - SemSeg \cite{narayanan19icra}       & 96.5           & 92.9      & 95.8           & 92.1          & 95.6          & 97.5          & 94.0          & 92.3 & 94.6          \\
MTAN \cite{liu19cvpr}       & 98.2           & 92.3      & 98.1           & 94.3          & 98.0          & 94.1          & 97.9          & 91.0 & 95.5          \\
Conv single, $\mathrm{CE}$ (ours)                   & \textbf{100.0} & 94.2           & 99.8 & 98.0 & 98.2 & 98.5 & 99.4 & 90.5          & 97.3 \\ 
Conv single, $\mathrm{CE}^{\text{R}}_{\text{MT}}$ (ours)                   & \textbf{100.0} & 94.6           & \textbf{100.0} & \textbf{98.6} & \textbf{98.5} & \textbf{98.9} & \textbf{99.8} & 91.2          & \textbf{97.7} \\ \hline
\end{tabular}%
}
\end{table}

We also ablate the impact of pre-training by comparing three different variants of our single-level model.
Table \ref{fm3m_our} shows that Vistas pre-training contributes more than using 
the larger ResNet-50 backbone.
Sequential enhancement is not applicable 
since this is a single-frame prediction task.

\begin{table}[h]
\centering
\caption{
  Ablation of pre-training on FM3m (AP): 
  H-\attribute{Highway}, R-\attribute{Road},
  Tu-\attribute{Tunnel}, E-\attribute{Exit},
  S-\attribute{Settlement}, O-\attribute{Overpass},
  B-\attribute{Booth}, Tr-\attribute{Traffic}.
}
\label{fm3m_our}
\resizebox{\columnwidth}{!}{%
\begin{tabular}{|l|c|ccccccccc|} 
\hline
\multicolumn{1}{|c|}{Model} & BB   & H              & R              & Tu             & E              & S              & O             & B              & Tr             & \uline{Mean}    \\ 
\hline
IN-1k (ours)          & rn18 & 99.2   & 91.5 & 99.6  & 97.3 & 97.8      & 97.0    & 98.2 & 89.9   & 96.3           \\
IN-1k (ours)          & rn50 & 99.6   & 92.5 & 99.9  & 98.2 & 98.0      & 97.4    & 98.9 & 90.5   & 96.9           \\
Vistas (ours)         & rn18 & \textbf{100.0} & \textbf{94.6} & \textbf{100.0} & \textbf{98.6} & \textbf{98.5} & \textbf{98.9} & \textbf{99.8} & \textbf{91.2} & \textbf{97.7}  \\
\hline
\end{tabular}
}
\end{table}

\subsection{BDD100k}

Table \ref{table:bdd100k} evaluates our method 
on the default and the cross-domain 
setup of 
the \attribute{Scene} task of BDD100k. 
Both setups include comparisons with 
Honda Frame - Mask, Honda Frame - SemSeg
\cite{narayanan19icra}, and 
MTAN \cite{liu19cvpr}.
The default setup also includes 
Local-Global FCRNN \cite{ni22tim},
which is a two-stream model 
that relies on local and global features 
of Faster RCNN and InceptionV2.
The cross-domain setup
also includes the "source-only" baseline from the
Sparse Adversarial Domain Adaptation (SADA)
paper \cite{saffari23tetci}.

\begin{table}[htb]
\centering
\caption{Comparison with prior work on the Scene task of BDD100k}
\label{table:bdd100k}
\resizebox{\columnwidth}{!}{%
\begin{tabular}{|lccc|}
\hline
\multicolumn{4}{|c|}{BDD100k default setup}                                                                             \\ \hline
\multicolumn{1}{|c|}{Model}                                         & \multicolumn{3}{c|}{Accuracy}            \\ \hline
\multicolumn{1}{|l|}{Honda Frame - Mask \cite{narayanan19icra}}                      & \multicolumn{3}{c|}{76.8}                         \\
\multicolumn{1}{|l|}{Honda Frame - SemSeg \cite{narayanan19icra}}                      & \multicolumn{3}{c|}{76.0}                         \\
\multicolumn{1}{|l|}{MTAN \cite{liu19cvpr}}                            & \multicolumn{3}{c|}{73.9}                         \\
\multicolumn{1}{|l|}{Local-Global FCRNN \cite{ni22tim}}                      & \multicolumn{3}{c|}{76.0}                         \\
\multicolumn{1}{|l|}{Conv single, $\mathrm{CE}$ (ours)}               & \multicolumn{3}{c|}{78.4}                \\ 
\multicolumn{1}{|l|}{Conv single, $\mathrm{CE}^\text{R}_{\text{MT}}$ (ours)}               & \multicolumn{3}{c|}{\textbf{78.7}}                \\ \hline
\multicolumn{4}{|c|}{BDD100k cross-domain setup}                                                                                      \\ \hline
\multicolumn{1}{|c|}{Model}                                                 & Cloudy           & Rainy          & Snowy         \\ \hline
\multicolumn{1}{|l|}{Honda Frame - Mask \cite{narayanan19icra}}                & 70.9               & 63.8              & 62.2           \\
\multicolumn{1}{|l|}{Honda Frame - SemSeg \cite{narayanan19icra}}                & 72.3               & 65.1              & 62.3           \\
\multicolumn{1}{|l|}{MTAN \cite{liu19cvpr}}                & 71.1               & 66.2              & 60.7           \\
\multicolumn{1}{|l|}{SADA \cite{saffari23tetci}}               & 70.5            & 62.7           & 59.1           \\
\multicolumn{1}{|l|}{Conv single, $\mathrm{CE}$ (ours)} & 75.9   & 71.4  & 70.0  \\ 
\multicolumn{1}{|l|}{Conv single, $\mathrm{CE}^\text{R}_{\text{MT}}$ (ours)} & \textbf{76.6}   & \textbf{71.5}  & \textbf{70.9}  \\ \hline
\end{tabular}%
}
\end{table}

The table shows that our single-frame models 
outperform all competing approaches. 
Our multi-task loss-weighting contributes 
in all experiments, 
while sequential enhancement is not applicable 
due to the single-frame prediction setup.
Note that all experiments
in the cross-domain section
train only on sunny images and report on  
cloudy, rainy, and snowy images. 


\section{Conclusion}
\label{sec:conclusion}

We have presented a two-stage approach 
for automatic collection of road-safety 
attributes in monocular video.
Our approach complements 
the baseline 
convolutional recognition 
in a local spatio-temporal context
with three contributions: 
semantic segmentation pre-training, 
multi-task recall-based loss weighting, 
and sequential enhancement. 

Our multi-task convolutional classifier 
consists of a shared backbone 
and per-attribute back-ends. 
We pre-train the backbone 
for semantic segmentation 
of street-level images 
as a part of an efficient 
dense prediction pipeline. 
Our baseline outperforms 
the state of the art on Honda scenes, 
which emphasizes the value 
of our subsequent two contributions. 
We address the extreme class imbalance 
with a multi-task variant
of recall-based loss weighting.
In this 
setup, 
the magnitudes of individual task losses are 
normalized in order 
to 
encourage commensurate contributions 
to the total loss. 
A closer look at iRAP attributes 
reveals that different attribute types
exhibit very different temporal behavior patterns.
For instance, intersections occur 
at discrete moments in time,
while the number of lanes
remains constant through many frames.
These considerations led us to 
correct the local predictions 
with attribute-specific recurrent 
models that learn the 
temporal behavior 
over a larger temporal context.

We have experimentally demonstrated a 
substantial impact of our contributions 
over the strong baseline. 
Each of our three contributions 
improve the performance 
across the board for all attributes. 
The combined approach outperforms the previous work 
on all tasks of the Honda Scenes dataset.
The greatest improvements occur
on the Road place task due to 
significant class imbalance and the 
availability of multi-frame input.
Our approach also delivers 
competitive performance on FM3m and BDD100k, 
in spite of a weaker backbone 
and inability to leverage sequential post-processing. 

We emphasize that our approach 
delivers competitive performance 
on a very low computational budget. 
The shared convolutional features are extracted 
by 
an efficient backbone only 
once per frame. 
Per-attribute recurrent models 
are extremely efficient due to low-dimensional inputs and outputs. 
Preliminary experiments have shown that 
a straightforward causal adaptation 
can deliver road-safety assessments in real-time 
even on a mobile device. 

To summarize, our main contributions are 
semantic pre-training,
multi-task recall-based loss weighting,
and sequential enhancement.
In addition, we introduce iRAP-BH 
as a novel dataset for visual road-safety assessment
that specifically focuses on iRAP attributes.
Future work should explore the sensitivity
of the proposed approach 
to various kinds of domain shift
through evaluation on novel datasets.
The new datasets should include multiple countries,
different types of cameras, various seasons, 
and challenging meteorological conditions. 
Moreover, it would be interesting 
to 
attempt further improvement 
by leveraging 
panoptic predictions 
and monocular reconstruction.

\bibliographystyle{IEEEtran}
\bibliography{IEEEabrv,mybibfile}

\newpage

\section{Biography Section}

\begin{IEEEbiography}[{\includegraphics[width=1in,height=1.25in,clip,keepaspectratio]{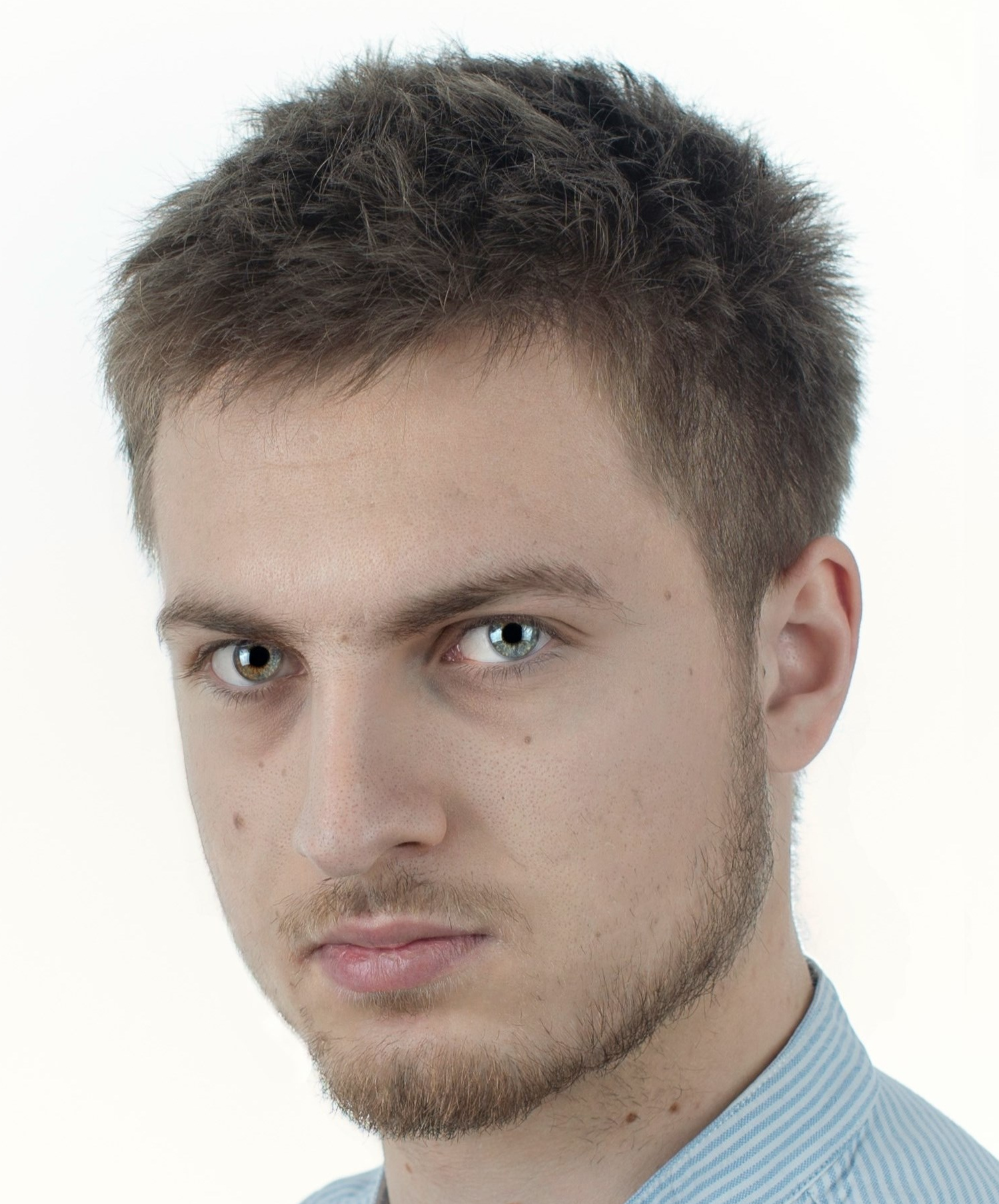}}]{Marin Kačan} received his MSc degree in computer science the University of Zagreb.
He spent two years as an expert associate at the Faculty of Traffic and Transport Sciences, University of Zagreb.
Currently, he is a young researcher and a PhD student at the Faculty of Electrical Engineering and Computing, University of Zagreb.
His research addresses image and video classification with a focus on road infrastructure assessment.
\end{IEEEbiography}


\begin{IEEEbiography}[{\includegraphics[width=1in,height=1.25in,clip,keepaspectratio]{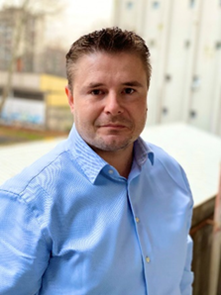}}]{Marko Ševrović} is a senior road safety engineer with more than 15 years of experience in the field of traffic flow theory, public transport planning, transport infrastructure design and road safety.  He holds a PhD degree in the field of traffic and transport technology from Faculty of Transport and Traffic Sciences, University of Zagreb, Croatia, where he works part-time as an Associate Professor and Head of Department of Transport Planning. In September 2018 he became part-time employee of the European Institute of Road Assessment (EuroRAP). He conducted and participated in more than 200 national and international professional and scientific projects in the field of road safety (EuroRAP/iRAP projects, Road Safety Audit), traffic planning and sustainable urban mobility planning. He is a charted engineer and road safety auditor.
\end{IEEEbiography}


\begin{IEEEbiography}[{\includegraphics[width=1in,height=1.25in,clip,keepaspectratio]{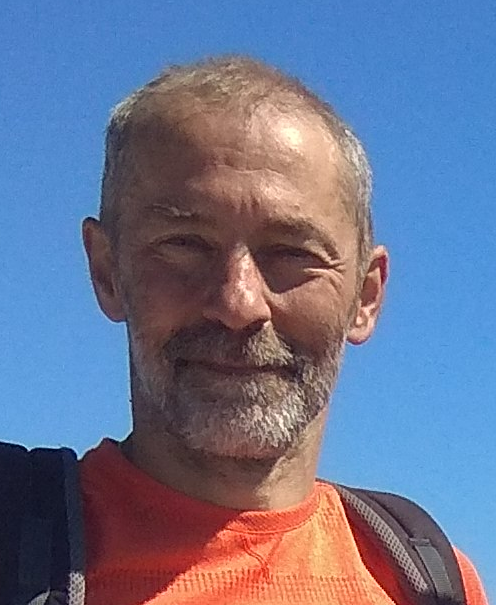}}]{Siniša Šegvić} received his Ph.D. degree in computer science from the University of Zagreb, Croatia. He was a post-doctoral Researcher at IRISA Rennes and TU Graz. 
He is currently leading a research laboratory 
at the Faculty of Electrical Engineering and Computing, University of Zagreb. 
His research involve 
computer vision and deep learning, 
with a special interest 
in robust scene understanding 
for autonomous vehicles and safe traffic.
\end{IEEEbiography}

\vfill
\clearpage

\end{document}